\documentclass[letterpaper, 10 pt, conference]{ieeeconf}  

\IEEEoverridecommandlockouts                              

\usepackage{graphicx}
\usepackage{comment}
\usepackage{amsmath,amssymb} 
\usepackage{color}
\usepackage{cite}

\usepackage{adjustbox}
\usepackage{multirow}

\title{\LARGE \bf
PBP-Net: Point Projection and Back-Projection Network for 3D Point Cloud Segmentation
}

\author{\authorblockN{JuYoung Yang\authorrefmark{1}
, Chanho Lee\authorrefmark{1}, Pyunghwan Ahn\authorrefmark{1}, Haeil Lee\authorrefmark{1}, Eojindl Yi\authorrefmark{1} and Junmo Kim\authorrefmark{1}}
\authorblockA{\authorrefmark{1}School of Electrical Engineering\\
Korea Advanced Institute of Science and Technology, Korea\\
Email: (yjy6711, yiwan99, dksvudghks, haeil.lee, djwld93, junmo.kim)@kaist.ac.kr}}

\begin{document}

\maketitle
\thispagestyle{empty}
\pagestyle{empty}

\begin{abstract}

Following considerable development in 3D scanning technologies, many studies have recently been proposed with various approaches for 3D vision tasks, including some methods that utilize 2D convolutional neural networks (CNNs). However, even though 2D CNNs have achieved high performance in many 2D vision tasks, existing works have not effectively applied them onto 3D vision tasks. In particular, segmentation has not been well studied because of the difficulty of dense prediction for each point, which requires rich feature representation. In this paper, we propose a simple and efficient architecture named point projection and back-projection network (PBP-Net), which leverages 2D CNNs for the 3D point cloud segmentation. 3 modules are introduced, each of which projects 3D point cloud onto 2D planes, extracts features using a 2D CNN backbone, and back-projects features onto the original 3D point cloud. To demonstrate effective 3D feature extraction using 2D CNN, we perform various experiments including comparison to recent methods. We analyze the proposed modules through ablation studies and perform experiments on object part segmentation (ShapeNet-Part dataset) and indoor scene semantic segmentation (S3DIS dataset). The experimental results show that proposed PBP-Net achieves comparable performance to existing state-of-the-art methods.

\end{abstract}

\section{Introduction}

Deep learning has shown successful results in various computer vision applications such as image classification, detection, and segmentation, owing to the development in 2D convolutional neural networks (CNNs)~\cite{krizhevsky2012imagenet,long2015fully}. 2D CNNs can efficiently capture global and spatial features with good generalization capability. Recently, improvement in 3D scanning devices have enabled the collection of 3D data. Thus, these devices have attracted significant attention for applications in several deep learning tasks such as 3D classification, object detection, and semantic segmentation. Previously, 3D data has been represented as multi-view images~\cite{su2015multi}, volumes~\cite{qi2016volumetric,li2016fpnn}, and structural data~\cite{klokov2017escape}. Compared to multi-view images and volumetric representations, point cloud, which refers to an unordered set of points, precisely represents data and requires lesser memory. 
Hence, a majority of recently proposed studies attempt to effectively deal with 3D point clouds.


\begin{figure}[t]
\begin{center}
        \centering
        \includegraphics[width=8cm]{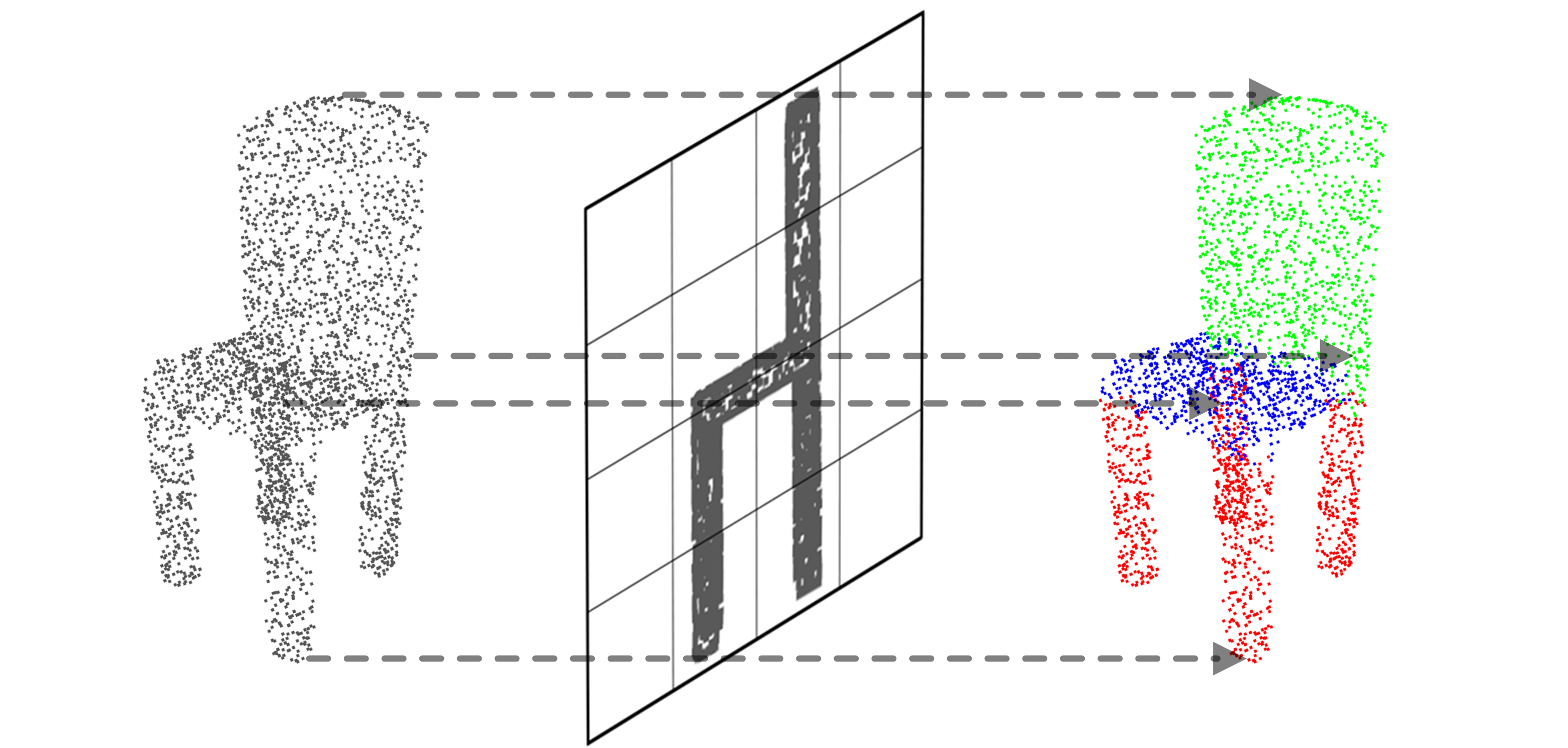}
    \caption{Brief concept of our idea. We project 3D points onto 2D planes to extract spatial features, and re-allocate these acquired features to the original points. We actually perform part segmentation for the chair and show the result}
    \label{fig:fig0}
\end{center}
\end{figure}

PointNet~\cite{qi2017pointnet} is a pioneering research for the classification and semantic segmentation of 3D point cloud. PointNet calculates the features of individual points via symmetric multi-layer perceptrons (MLP) to guarantee equivariance to the order of points. Previous studies have employed graph convolution, which is one of the most frequently used kernels to generate point-wise features, independent of the point order~\cite{wang2019dynamic,landrieu2018large}. Although these symmetric neural networks are capable of extracting local features of a point cloud, they often face difficulties when capturing the spatial features of point clouds. The most intuitive method to generate spatial features is directly using a 3D CNN through voxelization~\cite{papon2013voxel,maturana2015voxnet}. Even though 3D CNNs are capable of capturing the spatial features of high quality, these methods require significant computational capacities, which limits the size of the voxel and its scalability. Recent studies have demonstrated improved performance over these baseline methods; however, designing specialized neural networks for 3D point clouds is still a challenging task as compared to designing 2D CNNs.


As 2D CNNs exhibit good performances in various computer vision tasks, it is beneficial to utilize 2D CNNs for 3D recognition tasks. Even though several previous studies have applied 2D convolution for the classification~\cite{roveri2018network} and detection~\cite{lang2019pointpillars} from 3D point clouds, segmentation is more challenging task and requires dense predictions for each point. Also, most of their methods mainly focused on extracting global features or reasoning on 2D planes. On the other hand, we aim to extract discriminative features for each point. Some studies like~\cite{3DMVMVMV} utilize 2D CNNs by using voxel representations which requires huge memory allocation and limit its capacity. Since 2D CNNs have an effective performance in terms of memory and compuatation, we maximize that advantage by simplifying each process. 


In this paper, we propose a general framework of 2D CNN-based 3D point-wise feature extraction. The proposed method is computationally efficient and easy to implement. Our core idea is to first project 3D points and their local features onto 2D planes, and subsequently extracting spatial features by using 2D convolutional layers. Finally, these spatial features are back-projected onto 3D points. Figure~\ref{fig:fig0} briefly shows the concept of our idea. Based on this, we construct an efficient neural network termed as point projection and back-projection network(PBP-Net), which performs 3D point cloud segmentation tasks using 2D CNNs. Considering that 3D-to-2D projection and back-projection are differentiable, our PBP-Net is end-to-end trainable. Furthermore, PBP-Net achieves performance comparable to that of state-of-the-art methods and is applicable to various scales of datasets, including datasets with a large number of points or volumes, without any dramatic change in the network architecture.

The main contributions of this study are as follows:
\begin{enumerate}
    \item We suggest 3D-to-2D projection and 2D-to-3D back-projection modules to employ 2D CNNs as 3D point-wise feature extractors. Both modules are completely differentiable; hence, we can train the overall architecture end-to-end.
    \item PBP-Net is a simple yet powerful 2D convolution-based architecture for point cloud segmentation. Various 2D CNNs that are successful for 2D segmentation tasks can be directly utilized without any modifications. Thus, PBP-Net ensures structural freedom when choosing 2D convolutional backbones.
    \item The performance of PBP-Net is comparable to state-of-the-art methods, in terms of part segmentation and semantic segmentation. Our experimental results demonstrate that PBP-Net is capable of extracting 3D feature representations using 2D convolution.
\end{enumerate}


\section{Related Works}

In this section, we discuss the previous studies related to our method. As the proposed method aims to perform semantic segmentation of point clouds by using 2D CNNs, our work is related to studies pertaining to point cloud recognition as well as those pertaining to image segmentation. First, we introduce the deep learning-based methods for irregular 3D point cloud recognition. Subsequently, we review the convolution-based methods on regular grids in 2D (images) or 3D (shapes) spaces.

\textbf{Neural Networks on irregular point clouds.}
PointNet~\cite{qi2017pointnet}, which is a pioneering method for 3D point cloud recognition, acts as a simple baseline for various tasks such as classification and segmentation. PointNet employs MLP that extracts features from individual points. Thereafter, it uses the max-pooling features to produce global feature vectors. The global features can be used for object classification or they can be combined with point-wise features for segmentation (i.e., point-wise classification). PointNet achieves good performance for 3D object classification and part segmentation, and requires low computational capacities. However, follow-up studies have reported that PointNet has difficulties in capturing local features, which results from the learning interactions between neighboring points.

To extract better local features, previous works have attempted to generate local-to-global hierarchical features~\cite{qi2017pointnet++,hua2018pointwise} or to train the relationships between neighboring points, based on graph structures~\cite{wang2019dynamic,xu2018spidercnn,landrieu2018large}. PointNet++~\cite{qi2017pointnet++} was suggested for local feature extraction by applying PointNet to a set of points in the local neighborhood. Pointwise CNN~\cite{hua2018pointwise} has been proposed to voxelize the neighborhood of a point for the application of 3D convolution on the continuous space.

Previous studies have also attempted to apply graph neural networks for point cloud recognition. \cite{wang2019dynamic} proposed EdgeConv operator that effectively maps the relation between the neighboring features into edge weights such that the graph convolution can infer these relationships. 
SpiderCNN~\cite{xu2018spidercnn} proposed a convolution filter as a product of the step function and a Taylor polynomial to capture local geodesic information while maintaining the expressiveness of the kernel. In \cite{landrieu2018large}, the point cloud is initially over-segmented into simple parts of the objects, which are known as superpoints; subsequently, the relationships between these groups are trained via gated recurrent unit (GRU) structures.


\begin{figure*}[t]
\begin{center}
\includegraphics[width=17cm]{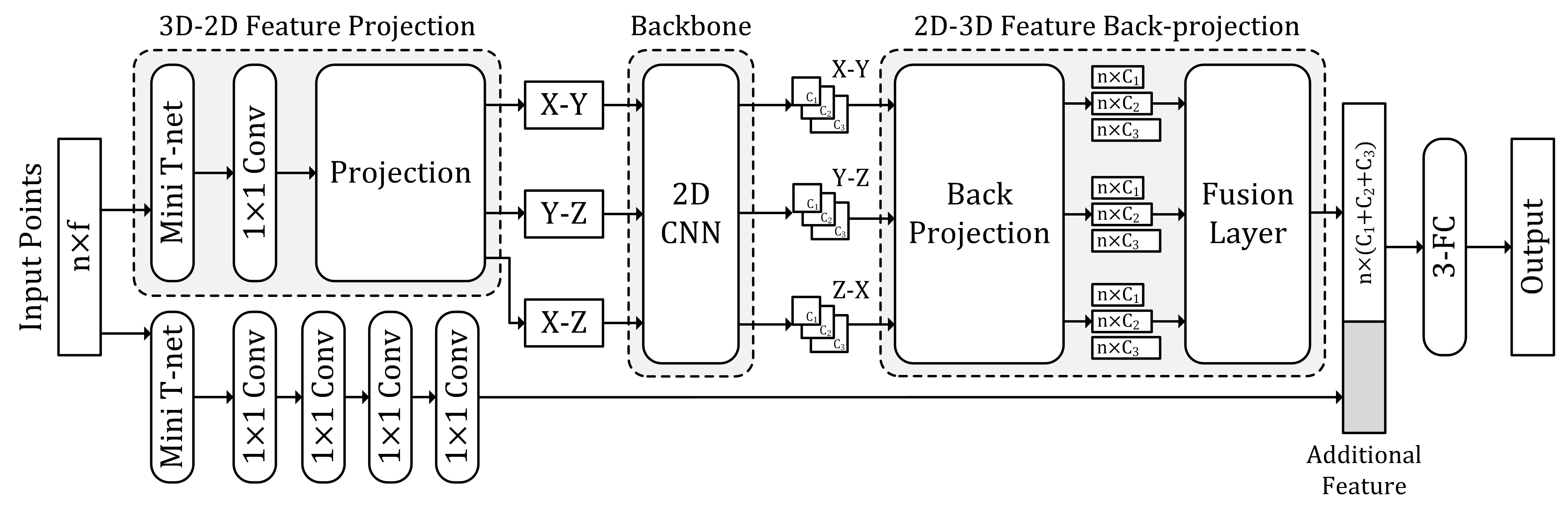}
\end{center}
\caption{Overview of our framework}
\label{fig:fig1}
\end{figure*}

\textbf{Neural Networks on regular grids.}
Convolutional neural networks can be used to perform 3D point cloud tasks in 2D or 3D spaces. To apply 3D convolution on point clouds, the point clouds must first be voxelized. In \cite{maturana2015voxnet}, voxels are generated as an occupancy grid with a binary value that indicates whether a voxel includes a point or not. However, 3D convolution requires significant computational resources, which prevents the voxelization of 3D spaces with high resolution. This limits the expressiveness of the voxelized representation of the point clouds, thereby limiting the performance. To overcome these problems, \cite{rethage2018fully} proposed the application of PointNet to a local neighborhood prior to using the 3D convolution layers. This alleviates information loss due to abstraction that arises when voxelizing the 3D space. 

Previous works have also aimed to transfer the 3D point cloud to the 2D domain and perform tasks using conventional 2D CNNs. \cite{su2015multi} proposed the use of a 2D CNN for the multi-view images of 3D shape models. This method extracts features from 2D images and then combines them using the view-pooling layer. \cite{roveri2018network} utilized a depth map to project point clouds onto a 2D plane and then applied a CNN to perform classification on the depth image. For 3D object detection, \cite{lang2019pointpillars} used pillar features, which were produced by applying PointNet at each pillar, as intermediate image features which were then processed via 2D convolution to produce the 3D bounding boxes.

Similarly, we also leverage conventional 2D image segmentation methods in our study. The fully convolutional network (FCN)~\cite{long2015fully} is one of the most widely used baseline for image segmentation. In the FCN, the last fully connected layers of the classification networks are replaced by (de)convolution layers such that pixel-wise predictions are produced. Following the principle of the FCN, \cite{noh2015learning} proposed a symmetric network structure wherein the first stage hierarchically encoded the features and the second stage applies deconvolution to produce pixel-wise classification results from the global features. \cite{ronneberger2015u} reported a similar structure with \cite{noh2015learning} and also combined intermediate features with the same spatial resolution at two stages. In this paper, we only report experimental results when using FCN~\cite{long2015fully} as a baseline. However, it should be noted that other network architectures such as \cite{noh2015learning,ronneberger2015u} can be used for 2D feature extraction and can possibly improve the performance.


\section{Method}

\subsection{Overview}

To demonstrate 3D feature extraction using 2D CNNs, we propose a framework that leverages 2D convolutional layers as a backbone. For input 3D point clouds, our architecture initially transforms them into 2D feature maps, then extracts hierarchical features using 2D convolutional layers, and finally projects the extracted features back onto 3D point locations. These three processes are separately explained in the following subsections. Our final architecture combines the proposed projection-based features with additional point-wise features to further improve performance, as explained in Section~\ref{sec:sec_3_3}. The overall architecture is presented in Figure~\ref{fig:fig1}.

\begin{figure}[t]
\begin{center}
        \centering
        \includegraphics[width=7.5cm]{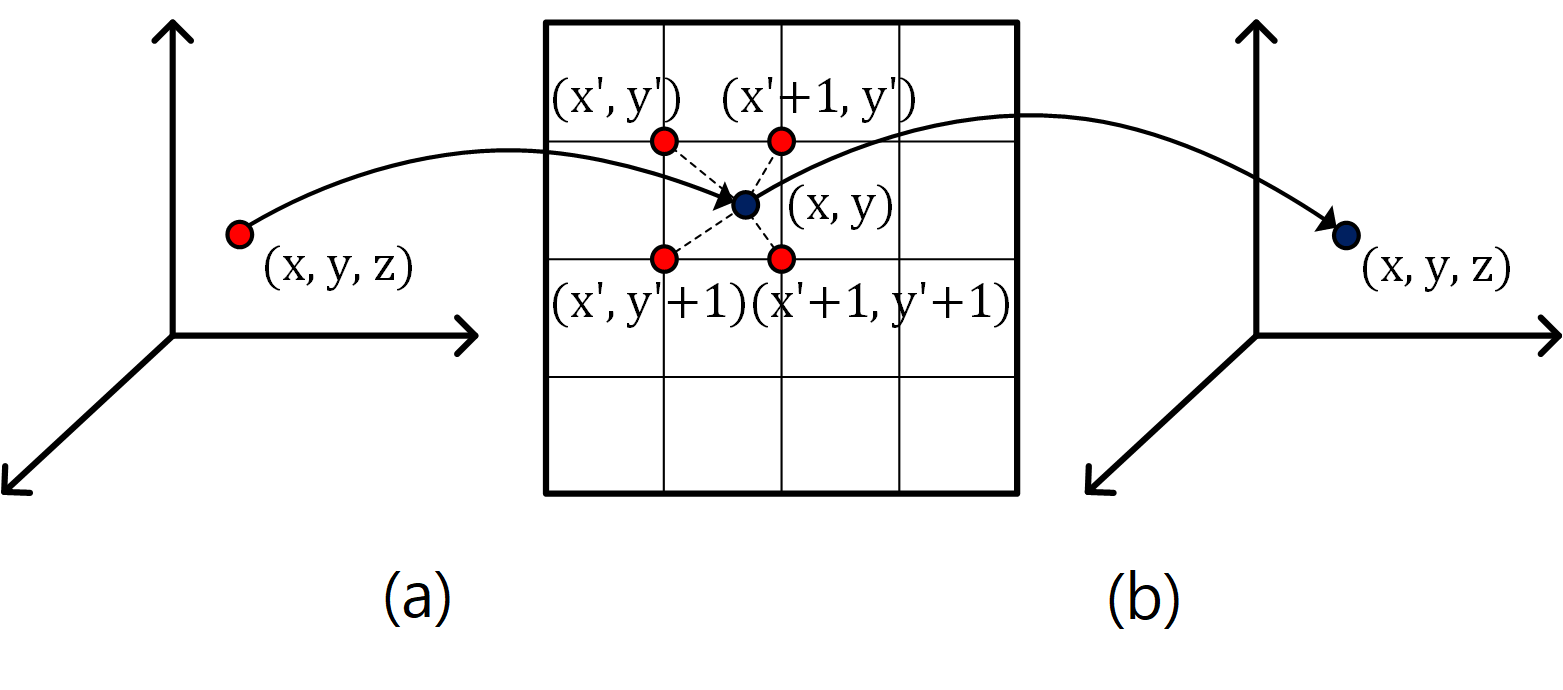}
    \caption{The operation of the (a) 3D-to-2D feature projection module and the (b) 2D-to-3D feature back projection is illustrated}
    \label{fig:fig2}
\end{center}
\end{figure}

\subsection{Proposed Modules}

\subsubsection{3D-to-2D Feature Projection}
\label{sec:sec_3_2_1}
The proposed architecture uses only point clouds as input, which are processed using 2D convolutional layers. However, as point clouds lie in a continuous 3D space, they cannot be directly fed into 2D CNNs. Hence, the point clouds need to be converted into 2D feature maps that lie on a regular grid, so that they can be processed by 2D CNNs. Alignment and shallow 3D feature extraction of the input point clouds are performed using mini T-net followed by a single convolutional layer. Mini T-net is a translation network that was used in Pointnet~\cite{qi2017pointnet}; it predicts a transformation matrix through a small network. We reduce the channel size of MLP layers to increase efficiency. Then, these point features are projected onto the XY-, YZ-, and ZX-planes. 


Let $p$ denote a single point in point clouds, with the coordinates $(x, y, z)$ and the feature $f$. First, we normalize the point cloud so that it lies in the given range for a plane. If the given plane is the XY-plane and the plane length is $l$, then $x$ and $y$ are normalized to be within $[-\frac{l}{2}, \frac{l}{2}]$. Therefore, we put $f$ on a grid on the XY-plane via bilinear interpolation. Although there are many interpolation methods that can be used to fill the grid, bilinear interpolation is adopted because of its low computation and memory requirement. If multiple features are assigned to the same cell of the grid, these features are added. This results in a 2D feature map on the $l*l$ grid that is transformed from a 3D point cloud. This transformation can be represented as the following equation:
\begin{equation}
I_{XY}(x',y') = \sum_{x}\sum_{y} G(x', y', x, y) \cdot f(x,y,z)
\end{equation}
\begin{equation}
G(A, B, a, b) = g(A, a) \cdot g(B, b)
\end{equation}
\begin{equation}
g(N, n) = \max (0, 1 - |N - n|)
\end{equation}

$I_{XY}(x',y')$ and $f(x,y,z)$ denote the 2D features on position $(x', y')$ of XY-plane and the 3D features of the point with coordinate $(x, y, z)$ respectively. G(·, ·) is the 2D bilinear
interpolation kernel that is composed of two one-dimensional kernels, as shown in Equation 2. We present Figure~\ref{fig:fig2} to help understand.

In the proposed method, three orthogonal planes are used for the projection rather than a single plane. The general projection and back-projection operations do not consider the information regarding the direction of projection; hence, using one or two planes does not guarantee that each point will possess distinctive features. Consider that there are two points $p$ and $p'$ with coordinates $(x, y, z)$ and $(x, y, z')$, respectively. When the XY-plane is used, projection and back-projection are performed by only using x and y values. Therefore, $p$ and $p'$ will possess identical features. To overcome this issue, we employ three non-parallel planes for the projection; in particular the orthogonal planes are used for simplicity. This allows each projection to extract different sub-features that complement each other, thereby ensuring that the proposed module is able to generate distinctive features for each point by using 2D convolutions. We demonstrate the effectiveness of using three orthogonal planes through comparative experiment, as described in Section~\ref{sec:sec_4_3}.


\begin{figure}[t]
\begin{center}
        \centering
        \includegraphics[width=5cm]{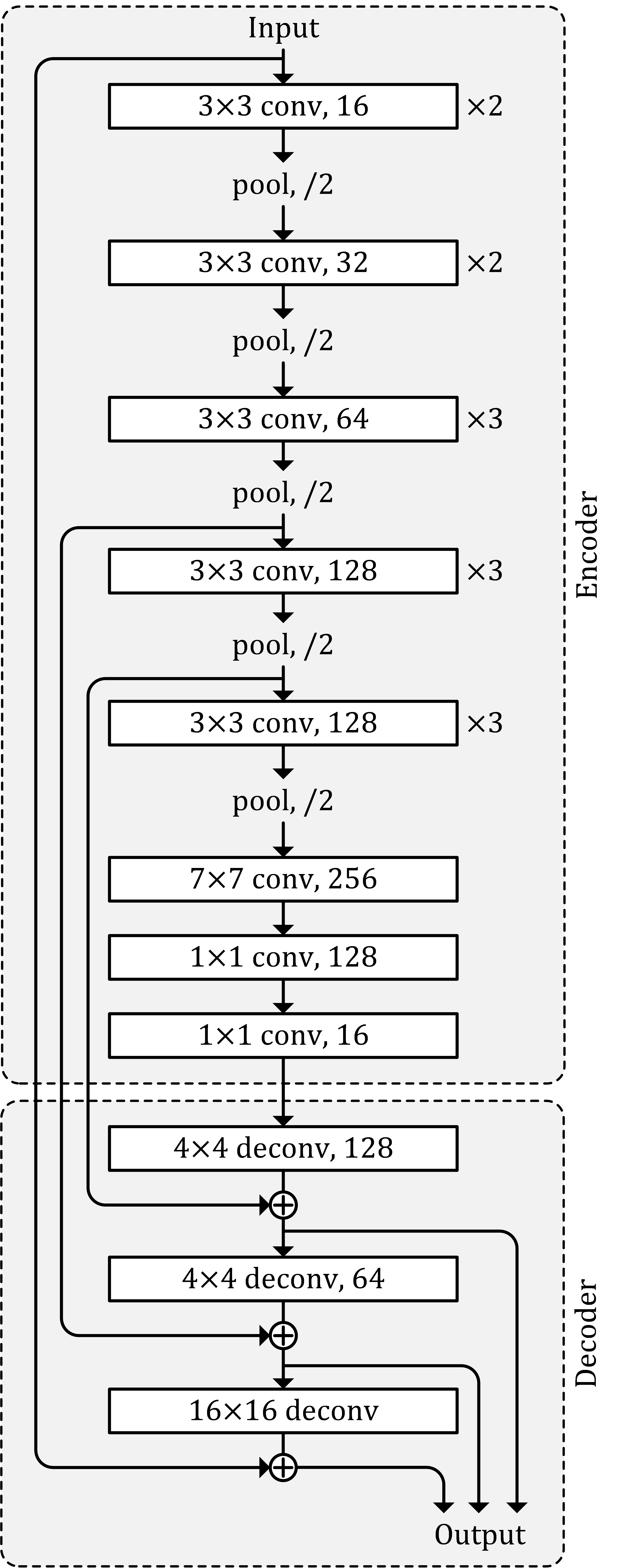}
    \caption{The architecture of the modified FCN-VGG16}
    \label{fig:fig3}
\end{center}
\end{figure}

\subsubsection{2D Convolutional Backbone}
\label{sec:sec_3_2_2}
The 2D features that are transformed by above-mentioned projection are inputted to the 2D convolutional backbone network to extract high-level features. We use the backbone network based on the FCN-VGG16~\cite{long2015fully}, which has been widely used for 2D segmentation tasks. However, we implement two modifications to improve efficiency and performance. First, we reduce the size of the channels in multiple layers by a factor of four to reduce the required memory and speed up the processing. Second, this network outputs feature maps from the intermediate layers as well as the final layer. This facilitates the creation of multi-scale features that possess rich information. We extract two intermediate feature maps that contain 128 and 64 channels and one final feature map that contains 16 channels from one input. As there are three inputs from the 3D-to-2D feature projection module, the total number of outputted feature maps is 9. This structure is presented in  Figure~\ref{fig:fig3}.


\subsubsection{2D-to-3D Feature Back-projection}
\label{sec:sec_3_2_3}
To apply the 2D features extracted using the 2D convolutional backbone to the point clouds, we back-project the 2D features to the 3D points. 
This is the reverse operation of the method described in Section~\ref{sec:sec_3_2_1}. We normalize $p$ to lie in the given image length range for a specific plane. Subsequently, we extract a new feature $f'$ from the grid by using bilinear interpolation. Therefore, in the case of the XY-plane, the transformation can be represented as follows:

\begin{equation}
f_{XY}(x,y,z) = \sum_{X}\sum_{Y} G(x, y, x', y') \cdot I_{XY}(x',y')
\end{equation}

$f_{XY}(x,y,z)$ and $I_{XY}(x',y')$ denote the 3D features of the point with coordinate $(x, y, z)$ from XY-plane and the 2D features at position $(x', y')$ of the XY-plane, respectively. G(·, ·) is the 2D bilinear
interpolation kernel, equivalent to that in Equation 2. The operation is shown in Figure~\ref{fig:fig2}.

After back-projection, we obtain 9 sub-features that were extracted from the 2D convolutional network for each point cloud. We combine these 9 sub-features by using the fusion layer to create a single integrated feature. Let each feature be $f_{P}^{d}$, wherein $P$ and $d$ are the given projection plane and the number of output dimension, respectively. 

The fusion layer first adds the sub-feature of different planes with the same dimension to create the intermediate feature $f^{d}$ for each dimension.

\begin{equation}
f^{128} = f_{XY}^{128} +  f_{YZ}^{128} +  f_{XZ}^{128}
\end{equation}
\begin{equation}
f^{64} = f_{XY}^{64} +  f_{YZ}^{64} +  f_{XZ}^{64}
\end{equation}
\begin{equation}
f^{16} = f_{XY}^{16} +  f_{YZ}^{16} +  f_{XZ}^{16}
\end{equation}

Subsequently, the integrated feature $f_{final}$ is created via channel-wise concatenate of the intermediate features

\begin{equation}
f_{final} = (f^{128},  f^{64},  f^{16})
\end{equation}


\subsection{Architecture and loss}
\label{sec:sec_3_3}
Our proposed architecture is simple and easy to implement. The overall architecture is composed of the three modules described in the previous subsection, followed by fully connected layers. For the final architecture, we add point-wise features to improve training. The additional features are simply extracted from the input point clouds using mini T-Net and fully connected layers. 
As every module performs differentiable operations, our framework is end-to-end trainable. For the loss function, simple point-wise cross-entropy is used for training. The point-wise cross-entropy function is the same as pixel-wise cross-entropy used in image segmentation:

\begin{equation}
CE(q,p)=-\sum_{n}\sum_{i} q_{n,i}\log(p_{n,i})
\end{equation}

where $p$ and $q$ denote the predictions and ground-truth labels, respectively; $n$ is the index of points, and $i$ refers to the class index.
%


\section{Experiments}
To demonstrate the effectiveness of 2D convolutions for 3D point cloud segmentation tasks, we perform extensive experiments. First, we evaluate our proposed architecture on various datasets and compare it to other state-of-the-art methods. Thereafter, ablation studies are performed to further investigate the proposed modules. 

\subsection{Datasets and Metrics}
This section summarizes the datasets and metrics used in experiments. We use S3DIS~\cite{s3dis} and ShapeNet-Part~\cite{shapenet} for segmentation task.

\begin{itemize}
	\item \textbf{S3DIS.} This dataset contains 3D scans of 6 indoor areas including 271 rooms. Each point is acquired from Matterport scanners and annotated with one of the 13 classes.
	\item \textbf{ShapeNet-Part.} This dataset contains 16,880 models-14,006 for training and 2,874 for testing. Each point is annotated with 2 to 6 parts. This dataset contains 16 classes with 50 different parts in total.
\end{itemize}



The mean intersection over union (mIoU) was used as the evaluation metric of the segmentation tasks. The IoU is computed as $\frac{TP}{TP+T-P}$, where $TP$, $T$, and $P$ are the number of true positive points, number of groundtruth points, and the number of predicted positive points. The mIoU is calculated by averaging the IoU of each class. 

\subsection{Implementation}
We implement the proposed architecture by using the Tensorflow framework~\cite{tensorflow}. The Adam optimizer is used and the size of each projection plane is set to (224, 224). 
To ensure that the sole performance of the 2D convolution operation could be analyzed, post processing techniques were not used. We emphasize that all the operations in the proposed architecture is implemented by using the built-in functions of Tensorflow and can be trained end-to-end from scratch.

\begin{itemize}
	\item \textbf{S3DIS.} We sample 4,096 points from the dataset and train the network for 50 epochs. The initial learning rate is 0.001, and the decay rate 0.5. The learning rate decay step is 300,000.
	\item \textbf{ShapeNet-Part.} 2,048 points are sampled from the dataset. We train the network for 200 epochs, and set the initial learning rate as 0.001. The learning rate is divided by 2 after 200,000 steps.
\end{itemize}

\begin{figure*}[t]
\begin{center}
\includegraphics[width=17cm]{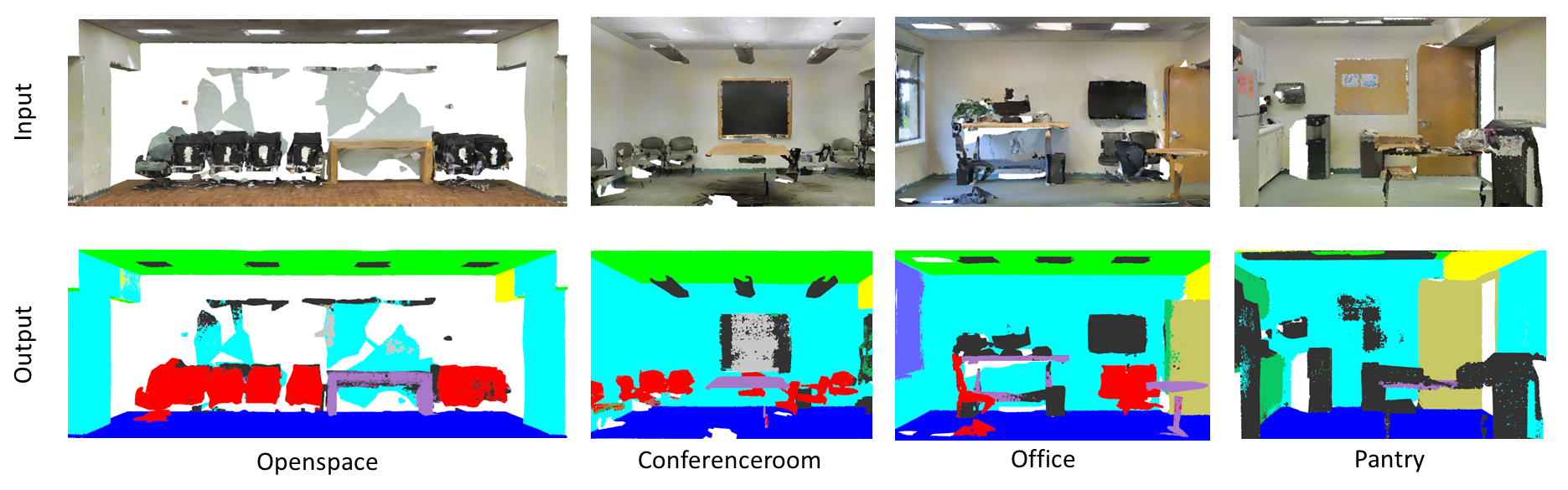}
\end{center}
\caption{Indoor scene semantic segmentation on the S3DIS dataset}
\label{fig:fig5}
\end{figure*}

\begin{table}[t]
\begin{center}
\caption{The results of the ablation studies on S3DIS Area5 dataset}
\begin{adjustbox}{max width=\linewidth}
\begin{tabular}{|c||c|cccc|}
\hline
Ablation & mIoU & \# Planes & T-Net & Multi-Scale & Additional\\
\hline\hline
\multirow{3}{*}{\# Planes} & 42.4 & 1 & & &\\
                           & 43.1 & 2 & & &\\
                           & 48.2 & 3 & & &\\

{T-Net} & 52.4 & 3 & \checkmark & &\\

{Multi-scale} & 53.2 & 3 & \checkmark & \checkmark &\\

{Additional} & \textbf{56.8} & 3 & \checkmark & \checkmark & \checkmark\\
\hline
\end{tabular}
\end{adjustbox}
\label{tab:tab1}
\end{center}
\end{table}

\subsection{Analysis}
\label{sec:sec_4_3}
We investigate the performance of different architectures to analyze our architecture through the ablation studies. As described in Section~\ref{sec:sec_3_2_1}, 
We compare the performance of the projection module by subtracting orthogonal planes. We also study the effect of the transformation network that exists in front of the projection module, multi-scale feature maps from the 2D backbone network, and additional features of the final architecture. All the experiments are performed as segmentation tasks using the S3DIS dataset. The S3DIS Area 5 dataset is used for testing, and the remaining areas datasets are used for training.
\medskip\\
\textbf{Projection of three orthogonal planes.} We vary the number of orthogonal planes on the projection module. Our experiments, which are listed in Table~\ref{tab:tab1} (\# Planes), prove that additional orthogonal planes facilitate the extraction of additional various features that complement each other. This is analogous to the approach used in multi-view-based studies~\cite{su2015multi,roveri2018network}. When the number of orthogonal planes in the projection are reduced, there is significant degradation in the performance. For two orthogonal planes and one orthogonal plane, the performance decreases by 5.1\% and 5.8\%, respectively.
\medskip\\
\textbf{Transformation Network.} Table~\ref{tab:tab1} (T-Net) lists the results of removing the transformation network that is in front of the projection module. We observe that the performance reduces by 4.2\%, which indicates that the transformation network assists in training by aligning the input coordinates and additional data.
\medskip\\
\textbf{Multi-scale feature maps.} We eliminated the multi-scale feature maps from the 2D backbone network. The result in Table~\ref{tab:tab1} (Multi-Scale) shows that the performance decreases by 0.8\%, which indicates that the multi-scale feature maps from the backbone network facilitate the creation of richer information by considering multi-scale features. This is inspired by the feature pyramid concept~\cite{fpn}, which is widely used for 2D segmentation applications.
\medskip\\
\textbf{Additional features.} The results in Table~\ref{tab:tab1} (Additional) shows that the final architecture, which is added additional features from the baseline architecture, accounts for an improvement of 3.6\%. This indicates that the additional features are well fused with the extracted features to assist in 3D feature extraction. It is noteworthy that even simple additional features such as those used in this study can improve performance.

\begin{table}[t]
\begin{center}
\caption{Object part segmentation scores on the ShapeNet-Part dataset (middle), and indoor scene semantic segmentation scores on the S3DIS dataset (right)}
\begin{tabular}{|ccc|}
\hline
Method & ShapeNet mcIoU & S3DIS mIoU\\
\hline\hline
SyncCNN~\cite{yi2017syncspeccnn} & 82.0 & -\\
SpiderCNN~\cite{xu2018spidercnn} & 81.7 & -\\
SplatNet~\cite{su2018splatnet} & \textbf{83.7} & -\\
SO-Net~\cite{li2018so} & 81.0 & -\\
SPGN~\cite{wang2018sgpn} & 82.8 & 50.4\\
KCNet~\cite{shen2018mining} & 82.2 & -\\
KdNet~\cite{klokov2017escape} & 77.4 & -\\
3DmFV-Net~\cite{ben20183dmfv} & 81.0 & -\\
DGCNN~\cite{wang2019dynamic} & 82.3 & 56.1\\
RSNet~\cite{huang2018recurrent} & 81.4 & 56.5\\
PointNet~\cite{qi2017pointnet} & 80.4 & 47.6\\
PointNet++~\cite{qi2017pointnet++} & 81.9 & -\\
SPG~\cite{landrieu2018large} & - & 62.1\\
PointConv~\cite{pointconv} & 82.8 & -\\
ResGCN-28~\cite{li2019deepgcns} & - & 60.0\\
\hline
ours & 82.7 & \textbf{62.8}\\
\hline
\end{tabular}
\label{tab:tab3}
\end{center}
\end{table}

\subsection{Comparison}
To verify the utility of 3D feature representations by using 2D convolution operations, we compare our final architecture to several state-of-the-art methods. 
We perform object part segmentation and semantic scene segmentation using ShapeNet-Part dataset and S3DIS dataset respectively. Unlike our previous ablation studies, we follow the 6-fold cross validation strategy for the S3DIS dataset. Table~\ref{tab:tab3} clearly indicate that our architecture is comparable to many existing methods, which means that 3D feature representation via the 2D convolution operation is effective, in terms of performance accuracy.


\begin{figure}[t]
\begin{center}
        \centering
        \includegraphics[width=8cm]{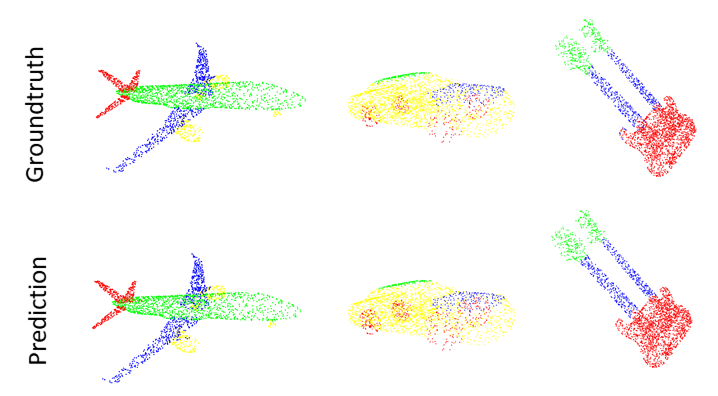}
    \caption{Object part segmentation on the ShapeNet-Part dataset}
    \label{fig:fig4}
\end{center}
\end{figure}

The results in Table~\ref{tab:tab3} (middle) show that the proposed architecture has a comparable performance with state-of-the-art methods, for object part segmentation task. 
Qualitative comparisons between our results and the groundtruth are depicted in Figure~\ref{fig:fig4}. We observe that the proposed method performs part segmentation robustly on many objects in the ShapeNet-Part dataset.

Table~\ref{tab:tab3} (right) shows the results of the indoor scene semantic segmentation task. Our architecture outperforms most state-of-the art methods, and in particular, we can see that it outperforms \cite{landrieu2018large} by 0.7\%. We also visualize our results and the groundtruth in Figure~\ref{fig:fig5}. 
It can be seen that the proposed architecture performs fairly well in general cases.


We demonstrate the comparable results with state-of-the art methods and believe that adjusting the hyper-parameters for training can result in further improvement.

\section{Conclusion}
In this paper, we propose PBP-Net, which is a simple yet effective neural network architecture leveraging 2D CNNs for 3D point cloud segmentation. In order to handle 3D point clouds, PBP-Net first projects 3D point cloud and their local features on multiple 2D planes. The features extracted by the 2D convolution backbone are re-allocated to their original point locations. We prove that a 2D convolution-based method can show comparable performance to that of other state-of-the-art methods.

PBP-Net provides a new perspective on the projection-based method as well as a baseline for using 2D CNNs for 3D tasks. Furthermore, the proposed method can be improved with the development of 2D segmentation, since we do not make any modification to the 2D CNNs when using them as backbone. 
We expect that this research will lead to the integration of 2D vision studies to 3D technologies, which have been remaining independent thus far.

\section*{Acknowledgements}
This research was supported by Naver Labs Corporation.

\end{document}